# Wheel Odometry-Based Localization for Autonomous Wheelchair

P Paryanto
*Department of Mechanical Engineering*
*Diponegoro University*
Semarang, Indonesia
paryanto@ft.undip.ac.id

Rakha Rahmadani Pratama
*Department of Mechanical Engineering*
*Diponegoro University*
Semarang, Indonesia
rakhardp@students.undip.ac.id

Roni Permana Saputra
*Research Centre for Smart Mechatronics*
*National Research and Innovation Agency*
Bandung, Indonesia
roni.permana.saputra@brin.go.id

*Abstract*— Localization is a fundamental requirement for an autonomous vehicle system. One of the most often used systems for autonomous vehicle localization is the global positioning system (GPS). Nevertheless, the functionality of GPS is strongly dependent on the availability of satellites, making it unreliable in some situations. As a result, autonomous vehicles must possess autonomous self-localization capabilities to ensure their independent operation. Odometry techniques are employed to achieve vehicle localization by predicting the vehicle position and orientation based on sensor measurements of the vehicle motion. One of the approaches employed in odometry is known as wheel odometry. Wheel odometry has a lower degree of reliance on the surrounding environment than visual odometry and laser odometry. This study aims to evaluate the performance of wheel odometry implementation for an autonomous wheelchair in the context of the localization process. The differential drive kinematic model is employed to determine the predicted pose of a wheelchair. This prediction is derived from the measurement of the linear and angular velocity of the wheelchair. Several experiments have been conducted to evaluate the performance of wheel odometry-based localization. Prior to experimenting, calibration procedures have also been performed to ensure accurate measurements of the sensor.

## I. Introduction

The field of assistive technologies has been significantly transformed in recent years due to the advances in robotics and artificial intelligence [1]-[5]. An example of an emerging assistive technology is the increasing availability of electric-powered personal mobility vehicles (PMVs) in numerous locations worldwide [6]. These technological advancements have also contributed to the rising popularity of autonomous driving systems (ADS) in mobility vehicles. As a result, there is a continuous effort to enhance the functionality and capabilities of ADS, with a primary focus on ensuring driver safety [7].

Localization is a crucial requirement in autonomous vehicles, as it facilitates the determination of the vehicle's position and orientation at regular intervals. Additionally, localization is vital in enabling navigation, object tracking, and obstacle avoidance capabilities. The most common form of localization in autonomous systems is the global positioning system (GPS). Nevertheless, even though the GPS can offer positioning information directly, its reliability still needs to be improved for the basic navigation system of an autonomous platform, as it solely relies on satellite signal strength [8].

Satellite signals are extremely susceptible to interference from various sources, such as specific objects or unfavorable weather conditions. These can block frequencies and interfere with signals sent by other electronic devices. A potentially dangerous situation for users could happen if an autonomous vehicle has trouble localizing or does not know exactly where it is [9]. This situation makes it hard for autonomous vehicles to decide what to do next. As a result, autonomous systems need a dependable localization system, which they can get by incorporating a different technique that ensures the system always has the localization data it needs for autonomous navigation.

One frequently used technique for achieving localization is using odometry methods, which involve estimating the movement and position of autonomous vehicles by leveraging measurements obtained from their movement sensors. The localization strategy based on odometry can be classified into several main categories, including wheel odometry, inertial odometry, visual odometry, laser odometry, and radar odometry [10].

The objective of this work is to include the wheel odometry technique into the current localization system of an autonomous wheelchair, which has been developed at the Research Centre for Smart Mechatronics, National Research and Innovation Agency, Indonesia. The wheelchair initially depended only on LiDAR and RGB-D sensors for its operations. LiDAR is used to identify and collect data on the distances between objects and surfaces. Meanwhile, the RGB-D camera exhibits great proficiency in accurate distance measurement and mapping initiatives due to its ability to gather depth data and build complex 3D point clouds. These sensors are utilized to perform simultaneous localization and mapping (SLAM) techniques.

However, it is important to note that both sensors have their limitations. For instance, the RGB-D camera is vulnerable to variations in lighting conditions, which might result in less precise and inconsistent outcomes [11]. LiDAR is susceptible to adverse weather conditions, including but not limited to direct sunlight and variations in object reflectance. Hence, leveraging wheel odometry becomes necessary to address this shortcoming.

Compared to visual and laser odometry, wheel odometry relies less on the surrounding environment. This odometry works even when cameras and LiDAR provide limited or no information [12]. The wheel odometry estimates the location relative to the starting point, and it determines the vehicle's linear displacement by counting the number of wheel spins. This odometry is a simple and low-cost localization approach that can be employed in vehicles [13]. This odometry is a non-holonomic system capable of interpreting forward, backward, and rotational motion [14]. Rotary encoder sensors can be utilized to implement the wheel odometry approach. This encoder is well-known for its capacity to transform wheel rotation into real-world movement data, promising enhanced accuracy and precision [9].

Nevertheless, wheel odometry exhibits some limitations, including inconsistencies in location estimation when operating on rough terrain compared to flat and smooth ground [10]. The utilization of RGB-D cameras and lidar technology can effectively address the limitations associated with the wheel encoder. Thus, integrating these three components is necessary to enhance the localization capabilities of autonomous vehicles.

This study focuses on utilizing the rotary encoder for the wheel odometry-based localization technique employed in autonomous wheelchairs. Incorporating wheel encoders, RGB-D cameras, and LIDAR sensors to improve localization capabilities falls beyond the scope of the present study. The main contribution of this work includes the design, modeling, and implementation of rotary encoders utilized for wheel odometry in the context of an autonomous electric wheelchair. Additionally, the study investigates the performance of localization based on proposed wheel odometry.

## II. METHOD

### A. Motion Model for the Differential Drive Vehicle

The use of odometry allows autonomous driving systems (ADS) to have their current location constantly monitored. The performance of wheel odometry in autonomous vehicles is dependent on the kinematic model that is utilized [15].

The majority of electric wheelchairs utilize a differential drive system, which consists of two wheels that are propelled along a common axis. By adjusting the speed of each wheel, it is possible to drive the vehicle in either a forward or backward direction, as well as directing it to the left or right. Motion controls are typically implemented via a joystick module in wheelchairs. The instantaneous centre of curvature (ICC) is a location along the left and right wheel axis where the vehicle must be when making a turn [16].

The coordinates of a vehicle with a two-wheeled differential drive system are computed depending on the speed of the wheels to give position and orientation information. In a Cartesian coordinate system, the vehicle position is commonly represented by the pair $(x, y)$, where $x$ is the horizontal position and $y$ is the vertical position. The angle $\theta$ to the reference axis represents orientation. Fig. 4 illustrates the two-wheeled differential drive system in Cartesian coordinates, where d is the wheel diameter, $L$ and $R$ are the left and right wheel distances, and $VL$ and $VR$ are the left and right wheel speeds.

The following equation can be employed to calculate the linear velocity $\dot{x}$ and $\dot{y}$, and angular velocity $\dot{\theta}$ of a vehicle given heading $\theta$, linear speed $v$, and angular speed $\omega$ in Cartesian coordinates:

$$\begin{aligned} \dot{x} &= v\cos(\theta) \\ \dot{y} &= v\sin(\theta) \\ \dot{\theta} &= \omega \end{aligned} \quad (1)$$

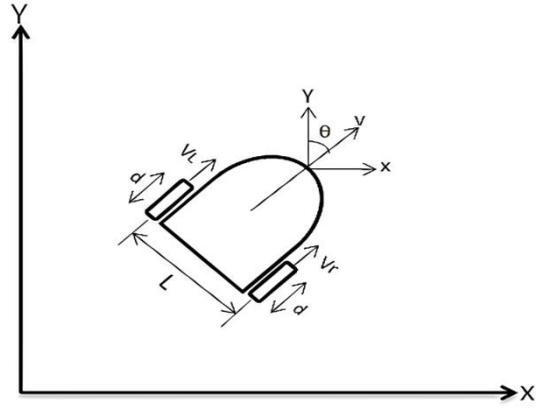

Fig. 1. The illustration model of a differential drive system in Cartesian coordinate [17].

Equation (1) may therefore be rewritten for a differential type of vehicle as:

$$\begin{aligned} \dot{x} &= \frac{R}{2}(V_r + V_l)\cos(\theta) \\ \dot{y} &= \frac{R}{2}(V_r + V_l)\sin(\theta) \\ \dot{\theta} &= \frac{R}{L}(V_r - V_l) \end{aligned} \quad (2)$$

If both wheels have the same speed, the resultant angular velocity is zero. When the vehicle goes in a straight x-axis direction, the value from the $X$ location grows, but the $Y$ value remains constant. On the other hand, when the vehicle angle is perpendicular to the x-axis, the value from the $Y$ location increases while the $X$ value remains constant. The displacement of the vehicle from the start to the end of the position determines the total movement of the vehicle.

### B. Wheel Odometry Approach for Predicting The Vehicle Pose

Odometry has associations with position and time, and it incorporates kinematics in the form of equations that correlate an object's location to time. In the context of one-dimensional motion, for instance, the position of an object can be represented by the function $x(t)$, where $x$ is the position and $t$ represents time. The odometry calculation can be done with the kinematic model based on the Taylor series expansion, as referenced in [18]. The expansion follows the following general form:

$$f(x) = \sum_{n=0}^{\infty} \frac{f^{(n)}(a)}{n!}(x-a)^n \quad (3)$$

Given the generic form in (3), the first order term of Taylor series expansion can be expressed as:

$$x(t + \Delta t) = x(t) + \dot{x}\Delta t \quad (4)$$

The following equation computes the odometry position of the vehicle using first-order Taylor series expansion:

$$\begin{aligned} x_{k+1} &= x_k + D_c \cos(\theta) \\ y_{k+1} &= y_k + D_c \sin(\theta) \\ \theta_{k+1} &= \theta_k + \frac{D_r - D_l}{L} \end{aligned} \quad (5)$$

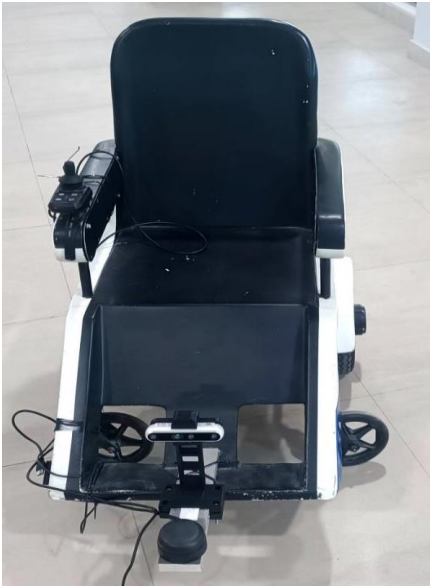

Fig. 2. Electric wheelchair platform used for the proposed wheel odometry implementation.

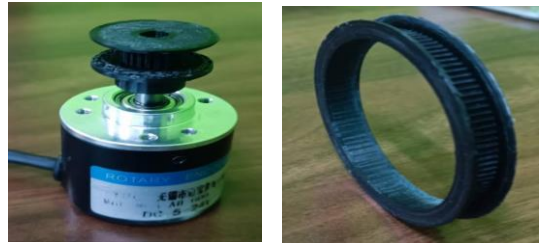

Fig. 3. The 3D printed pulleys (rotary encoder and wheel pulleys) used in the setup presented in this paper.

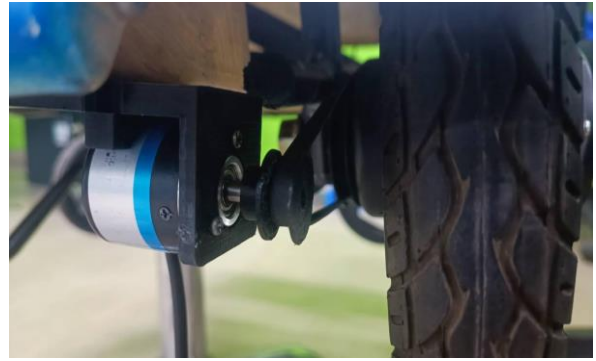

Fig. 4. The integration of the rotary encoders for wheel speed sensor in the wheelchair platform.

The variable $k$ represents the present location of the vehicle during a specified sampling interval $\Delta t$, $D_c$, $D_l$, and $D_r$. This position is calculated as the change in displacement of the right and left wheels.

In the context of odometry, the process involves frequent calculation and updating of the vehicle's position and movement parameters. This is achieved by determining the distance travelled and changes in orientation at regular intervals, in which the 1000 millisecond time interval is used in this work. These calculations are performed to update the estimated position and direction of the vehicle continually. If this condition is met, it will update the previous timestamp to the current time and proceed with the same procedure for the subsequent period. The process facilitates the maintenance of precise odometry data while guaranteeing a regulated speed.

In order to provide desired outcomes, an enhancement was made to the wheel odometry approach by incorporating a filter that encompasses rpm calibration through the utilization of a tachometer, as well as linear speed and angular speed filters. The calibration of odometry plays a crucial role in mitigating the propagation of errors [19].

### III. System Design and Implementations

#### A. Mechanical Design

Designing an efficient odometry system based on mechanics poses a formidable engineering challenge, particularly due to the inherent complexity of real-world environments. In the realm of autonomous mobility, where robots, including autonomous wheelchairs, operate in diverse and unpredictable terrains, the need for accurate localization becomes essential. The efficiency of an odometry system is profoundly influenced by unmodeled effects, especially in uncertain terrain conditions [20].

The wheelchair platform used in this study is equipped with wheel odometry. Each driving wheel of the wheelchair is powered by a brushless DC (BLDC) motor. The motors possess the following comprehensive specifications: operating at a voltage of 24 volts, generating a power output of 150 watts, and capable of delivering a maximum torque of 30 N.m. Additionally, each motor weighs 2.9 kg, rotates at a speed of 106 revolutions per minute (rpm), and is designed to fit a wheel with a diameter of 30 cm. Furthermore, the motor is equipped with a single axle and has an electromagnetic brake (EMB) feature. The combined mass of the wheelchair amounts to 62 kg.

Wheel speed sensors are required to establish the wheelchair's location using odometry. The rotary encoders, utilized as the speed sensors, is a 2-phase incremental rotary encoder with a pulse count of 600 and a working voltage range of 5 to 24 volts. In order to integrate the rotary encoder with the wheelchair wheel, pulleys and a timing belt transmission system that connects the wheel and the rotary encoder is used. The pulley is attached to the shaft of the BLDC motor and the rotary encoder. Both pulleys are linked via a timing belt to transmit the wheel rotation into the rotary encoder.

The timing belt utilized in this application has a pitch of 2mm, while the pulleys used consist of a 20-tooth pulley and a 100-tooth pulley. The rotary encoder is connected to the 20-tooth pulley, while the wheel is connected to the 100-tooth pulley. With this setting, the ratio of this transmission is 1:5. Fig. 3 shows the 3D printed pulleys used in this setup.

In order to attach the rotary encoder to the wheelchair platform, a holder for the rotary encoder was made. The holder is then affixed to the body under the wheelchair so that the timing belt can link the rotary encoder pulley and the wheel pulley. Fig. 4 shows the integration of the rotary encoders, pulleys, and timing belt on the wheelchair platform.

#### B. Electrical Design

Despite the fact that the real wheelchair platform has an RGB-D camera and a LIDAR sensor, we focuses on developing a wheel odometry system for localization and evaluating the performance of the proposed implementation in this study. Fig. 5 illustrates the schematic representation of the odometry system implemented in this study.

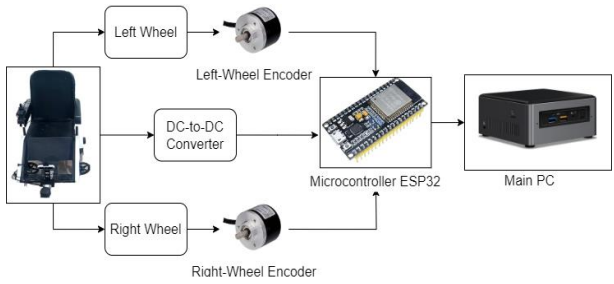

Fig. 5. Block diagram of the schematic representation of the odometry system implemented in this study.

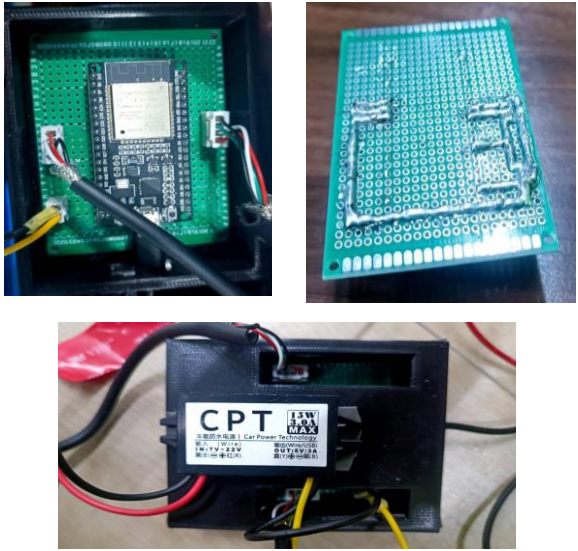

Fig. 6. The printed circuit board (PCB) implementation of the diagram illustrated in Fig. 5, and the circuit cover manufactured utilizing 3D printing technology.

As can be found on the diagram, the signals from the left-wheel and right-wheel rotary encoders are read by an ESP32 microcontroller module. The signals are then processed to calculate the rotational speeds of the right and left wheels of the wheelchair. The microcontroller is powered by the wheelchair's 24 V battery, which is converted to 5 V using a step-down DC-to-DC converter. The encoder signal result data is transmitted to the main computer via USB to serial connection.

The printed circuit board (PCB) implementation of the diagram demonstrated in Fig. 5 could be found in Fig. 6. Additionally, a circuit cover is manufactured utilizing 3D printing technology, taking into account both appearance and functionality.

## IV. EXPERIMENT & RESULTS

A calibration was performed initially on the rotary encoder that was mounted on the wheelchair before any experiments were carried out. This was done by comparing the findings of the rpm value from the calculation results from the microcontroller module with the actual wheel rotational velocity measured by a laser-based tachometer.

TABLE I. COMPARISON OF ROTARY ENCODER READINGS WITH TACHOMETER.

| Tachometer (rpm) | Right Encoder (rpm) | Left Encoder (rpm) |
|---|---|---|
| 33 | 33 | 33 |
| 53 | 54 | 55 |
| 72 | 74 | 75 |
| 92 | 94 | 95 |
| 114 | 118 | 118 |
| 137 | 141 | 142 |

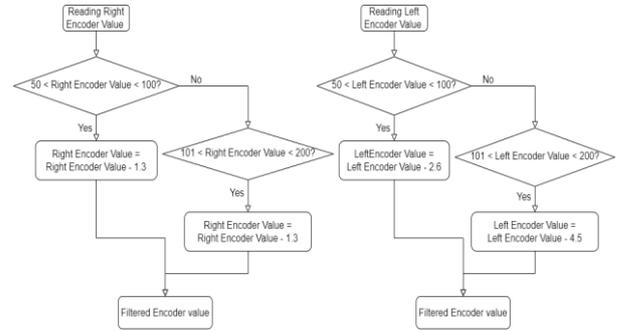

Fig. 7. The calibration process for minimizing the encoder reading error.

Table 1 shows the rotary encoder calibration results. According to Table 1, an inaccuracy is seen when the number of rotations per minute (rpm) exceeds a value of 50. An issue arises within the rotary encoder, resulting in distinct errors for each speed. As the speed increases, the error also increases. Therefore, we conducted the calibration process to minimise the encoder reading error using the filtering method as illustrated in Fig. 7.

The calibration filter is implemented to account for the discrepancy in rotary encoder data between the two wheels. While complete accuracy cannot be guaranteed, it has the potential to reduce errors to a certain degree. Subsequently, it is important to calibrate the linear velocity discrepancy between the two wheels in order to assess any variations shown by the wheelchair during linear motion at a rate of radiation per second. Table 2 shows the discrepancy rotational speed data between the right-wheel encoder and left-wheel encoder in radian per second.

The angular speed of the two wheels has a maximum difference of up to 0.1 rad/s. This difference is due to deviation or what is called drift in wheelchairs. Therefore, we implemented filtering data approach as illustrated in Fig. 8.

TABLE II. COMPARISON OF ROTARY ENCODER IN RADIAN PER SECOND BETWEEN RIGHT-WHEEL ENCODER AND LEFT-WHEEL ENCODER.

| Right Wheel (rad/s) | Left Wheel (rad/s) |
|---|---|
| 0,15 | 0,15 |
| 0,66 | 0,64 |
| 1,33 | 1,34 |
| 1,52 | 1,57 |
| 1,63 | 1,66 |
| 1,81 | 1,71 |

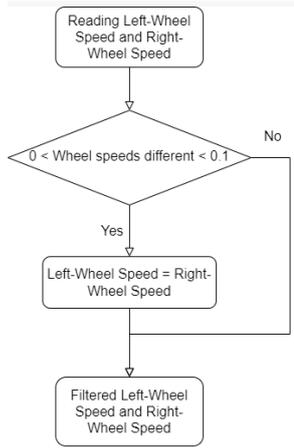

Fig. 8. Filtering process for the left-wheel speed and right-wheel speed reading.

Furthermore, a comparative analysis was also conducted to compare between the actual heading angle of the wheelchair and the estimated heading angle obtained from the differential kinematics equations processed from the rotary encoder data. We conducted the experiments by turning the vehicle to actual heading angle of 90 degrees and comparing it with the estimated heading angle.

After conducting a series of experiments involving the rotation of a wheelchair by 90 degrees, a set of data was collected. The recorded values for each trial were as follows: 20, 22, 30, 32, and 36. To account for the variations seen in the data, a filtering process was implemented by calculating the average. The resulted average data is then divided by the actual heading angle of the wheelchair (i.e., 90 degrees). The results shows that the estimated heading angle could be calibrated by a proportional gain of 3.2.

After obtaining the calibration factor, the evaluation was carried out by replicating the preceding heading angle comparison experiments while integrating the obtained calibration factor. The outcomes demonstrate that the estimated heading angles derived from the differential kinematics equations exhibit a substantial enhancement through the attainment of a considerably reduced maximum error of approximately 6 degrees. Considering the obtained information, another filtering process, as illustrated in Fig. 9, is then carried out prior to the final experimental evaluation. Furthermore, it is important to mention that angle wrapping will take place at a heading angle of 366 degrees once the filtering procedure completes one complete rotation.

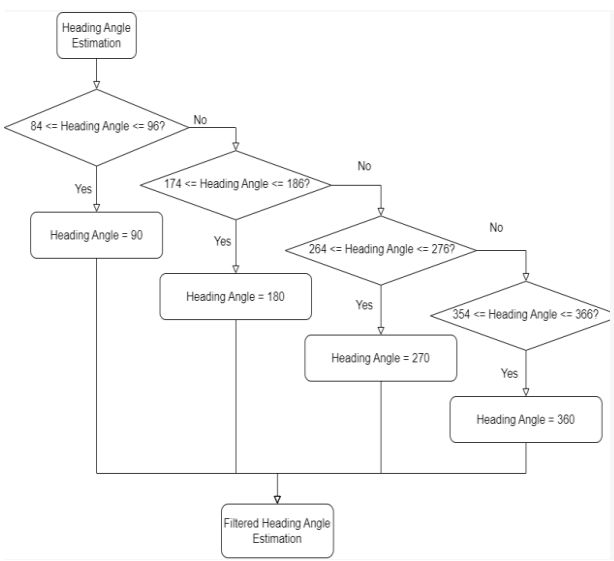

Fig. 9. Filtering process for the heading angle estimated based on the sensor reading.

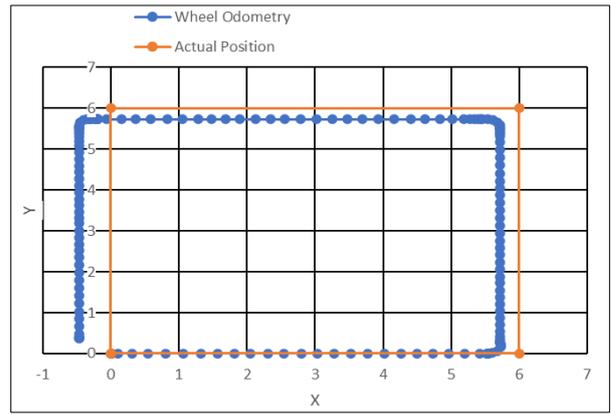

Fig. 10. The experimental results of wheel odometry-based localization implemented on the electric wheelchair platform. The results illustrate the comparison between the actual path and the estimated path generated from the proposed wheel odometry system.

TABLE III. COMPARISON OF THE WHEELCHAIR POSITION COORDINATES ESTIMATED BY THE WHEEL ODOMETRY WITH ACTUAL COORDINATES

| Wheel Odometry | | Actual Position | |
|---|---|---|---|
| Position X | Position Y | Position X2 | Position Y2 |
| 0 | 0 | 0 | 0 |
| 5,55 | 0 | 6 | 0 |
| 5,55 | 5,5 | 6 | 6 |
| -0,47 | 5,5 | 0 | 6 |
| -0,47 | 0,38 | 0 | 0 |

Once all the necessary filtering procedures have been completed, the experiment is prepared for implementation. The experiments were carried out on a 6x6 meter square path with the wheelchair regulated to follow the predetermined square path. Throughout the operation along the path, the estimated heading and real-time position derived from the differential kinematics equations and processed from the rotary encoder data were then collected. The estimated data was later compared to the predefined actual path. Fig. 10 shows the comparison between the actual path and the estimated path generated from the proposed wheel odometry system.

Fig. 10 and Table 3 shows that upon the wheelchair's completion of a 6x6 meter square trajectory and return to the beginning coordinates (0.0), there is an observed discrepancy in the wheel odometry. The error in the x coordinate is measured to be 0.47 m, while the error in the Y position is measured to be 0.38 m. These errors are assessed in relation to the displacement of the wheelchair at the completion of the task, compared to the actual position. More specifically, the spatial location of each coordinate exhibits variation; for instance, when traversing towards the coordinates (6,6), the odometry readings indicate the coordinates as (5.55, 5.5). The discrepancy is also observed in the error value of the odometer reading across many coordinates.

These results demonstrates that even though several filtering and calibration processes have been carried out to improve the wheel odometry accuracy, the nature of odometry drift cannot be completely eliminated. Thus, integrating data from multiple sensors of different modalities (i.e., sensor fusion) is highly favoured to enhance the accuracy, reliability, and robustness of state estimation of an autonomous system. Multimodal sensor fusion increases not just the overall accuracy of state estimation but also its reliability and robustness in real-world, dynamic environments by using the complementing capabilities of diverse sensors.

## V. Conclusion

This study presents a comprehensive investigation of the modelling, design, and implementation of a localization system intended for an autonomous wheelchair, with a particular emphasis on wheel odometry. More specifically, this study investigates the effectiveness of localization by employing the proposed wheel odometry technique.

Based on the findings of the experimental study conducted using wheel odometry on a 6x6 meter square track, it was observed that there is a discrepancy of 0.47 m in the x-coordinate and 0.38 in the y-coordinate when the wheelchair returns to its initial starting point. Furthermore, the results consistently indicated coordinates of (6,6) while the wheel odometry recorded (5.55, 5.5). These outcomes were achieved after subjecting the data to various filters, including rpm adjustments using a tachometer, angular speed corrections, and heading angle estimation adjustments.

Wheel odometry is useful for localization and navigation in robotics, especially autonomous wheelchairs. This system poses various strengths, including cost-effectiveness, low power consumption, real-time data, simplicity and simplicity of implementation, and high update rates. Despite its advantages, it is critical to recognize that wheel odometry has several limitations, particularly in difficult conditions with uneven terrain or wheel slippage. As a result, it is frequently combined with additional sensors to improve the overall accuracy and resilience of the localization system.

## Acknowledgment

The experiments conducted in this study are fully funded by the Research Organization for Electronics and Informatics, National Research and Innovation Agency (Organisasi Riset Elektronika dan Informatika, Badan Riset dan Inovasi Nasional - BRIN). The study is a collaborative effort between the National Research and Innovation Agency and Diponegoro University.